\documentclass{amia}
\usepackage{lipsum} 

\usepackage{inconsolata}

\usepackage{microtype}

\usepackage{multirow}
\usepackage{multicol}
\usepackage{booktabs}
\usepackage{graphicx}
\usepackage{caption}
\usepackage{subcaption}
\usepackage[inline]{enumitem}
\usepackage{amsmath}
\usepackage[superscript]{cite}
\usepackage{subcaption}

\usepackage[table]{xcolor}

\usepackage{hyperref}

\DeclareCaptionFormat{boldnumber}{\textbf{#1#2}#3}
\begin{document}

\title{Large Language Models Struggle in Token-Level Clinical Named Entity Recognition}

\author{Qiuhao Lu, Ph.D., Rui Li, Ph.D., Andrew Wen, M.S., Jinlian Wang, Ph.D., Liwei Wang, M.D., Ph.D., Hongfang Liu, Ph.D.}

\institutes{McWilliams School of Biomedical Informatics, University of Texas Health Science Center, Houston, TX, USA
}


\maketitle

\section*{Abstract}
\textit{Large Language Models (LLMs) have revolutionized various sectors, including healthcare where they are employed in diverse applications. Their utility is particularly significant in the context of rare diseases, where data scarcity, complexity, and specificity pose considerable challenges. In the clinical domain, Named Entity Recognition (NER) stands out as an essential task and it plays a crucial role in extracting relevant information from clinical texts. Despite the promise of LLMs, current research mostly concentrates on document-level NER, identifying entities in a more general context across entire documents, without extracting their precise location. Additionally, efforts have been directed towards adapting ChatGPT for token-level NER. However, there is a significant research gap when it comes to employing token-level NER for clinical texts, especially with the use of local open-source LLMs. This study aims to bridge this gap by investigating the effectiveness of both proprietary and local LLMs in token-level clinical NER. Essentially, we delve into the capabilities of these models through a series of experiments involving zero-shot prompting, few-shot prompting, retrieval-augmented generation (RAG), and instruction-fine-tuning. Our exploration reveals the inherent challenges LLMs face in token-level NER, particularly in the context of rare diseases, and suggests possible improvements for their application in healthcare. This research contributes to narrowing a significant gap in healthcare informatics and offers insights that could lead to a more refined application of LLMs in the healthcare sector.}

\section*{Introduction}

Electronic Health Records (EHRs) are a key component in modern healthcare, encapsulating vast amounts of patient data, most notably in clinical notes. Their widespread adoption by healthcare providers in recent years has revolutionized the manner in which patients' visits and health information are recorded and managed \cite{henry2016adoption}\hspace{-0.1em}, thereby elevating the quality of patient care. However, extracting pertinent information from these records presents a significant challenge due to the abundant, heterogeneous, and private nature of clinical texts. At the heart of addressing this challenge is Named Entity Recognition (NER), a fundamental task in natural language processing aimed at identifying and categorizing key information units in the text. In the clinical domain, the goal of the task is to identify all occurrences of specific clinically relevant named entities in the given unstructured clinical narrative \cite{kundeti2016clinical}\hspace{-0.1em}.

There has been a surge of interest in developing clinical NER systems in the last few years. Early methods mostly rely on manually-crated rules and traditional machine learning techniques, such as MetaMap \cite{aronson2010overview}\hspace{-0.1em}, KnowledgeMap \cite{denny2003knowledgemap}\hspace{-0.1em}, cTAKES \cite{savova2010mayo}\hspace{-0.1em}, etc. With the trending of deep learning methods and the Transformer architecture \cite{vaswani2017attention}\hspace{-0.1em}, more researchers shift to building clinical NER systems and other NLP applications upon pre-trained language models such as BERT \cite{devlin2019bert}\hspace{-0.1em}. A typical solution is to insert a multilayer perception (MLP) on top of the language model and train the entire model via fine-tuning. For example, Alsentzer \textit{et al.} fine-tune BioClinicalBERT on four i2b2 NER tasks \cite{alsentzer-etal-2019-publicly}\hspace{-0.1em}. Similarly, Sung \textit{et al.} present BERN2, which uses Bio-LM \cite{lewis-etal-2020-pretrained}\hspace{-0.1em} as the foundation model and achieves better performance via multitask learning \cite{sung2022bern2}\hspace{-0.1em}.

A common theme among the aforementioned deep-learning-based methods is their data-hungry nature, which unfortunately poses significant challenges in the medical field, where issues of data scarcity, heterogeneity, and confidentiality are consistently prevalent. The situation is even worse for \textit{rare diseases} due to their diversity, complexity, and specificity. Rare diseases refer to diseases or conditions that affect fewer than $200,000$ Americans by definition in the United States \cite{dharssi2017review}\hspace{-0.1em}. These conditions are often underrepresented in datasets, leading to a scarcity of information that deep learning models can learn from.

Recent advancements in computational linguistics have led to the emergence of Large Language Models (LLMs). These models have shown remarkable capabilities in various domains, including the clinical field \cite{thirunavukarasu2023large}\hspace{-0.1em}. Their strengths in in-context learning, such as retrieval-augmented generation (RAG), present new opportunities for tackling the challenges of NER in clinical texts. Despite the promise of LLMs, most existing research in this area has either focused on document-level NER (i.e., aggregation NER), which involves identifying entities at a broader level in entire documents or paragraphs without predicting the exact span or location of these entities, or on adapting ChatGPT to this task. For example, Zhou \textit{et al.} introduce UniversalNER, a distillation approach using mission-focused instruction-tuning to create efficient models that excel in NER. They distill ChatGPT into a more compact LLaMA-based \cite{touvron2023llama}\hspace{-0.1em} model UniversalNER and demonstrate superior document-level NER accuracy across diverse domains, including biomedicine \cite{zhou2023universalner}\hspace{-0.1em}. Similarly, in their exploration of ChatGPT-3.5-turbo, Shry \textit{et al.} focus on the extraction of rare diseases and their associated phenotypes via prompt engineering, also limiting their analysis to document-level output, which overlooks the precise span and location of these entities \cite{shyr2024identifying}\hspace{-0.1em}. In contrast, the potential of LLMs, especially local open-source LLMs, in token-level NER (i.e., span-based NER) in clinical settings, particularly for rare diseases, remains relatively unexplored. 

\begin{figure}[t]
    \centering
    \begin{subfigure}[b]{0.2\textwidth}
        \includegraphics[width=\textwidth]{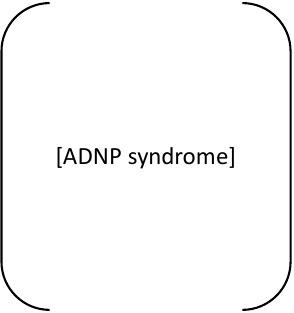}
        \caption{document-level}
        \label{fig:first_subfig}
    \end{subfigure}
    \hspace*{30pt}
    \begin{subfigure}[b]{0.2\textwidth}
        \includegraphics[width=\textwidth]{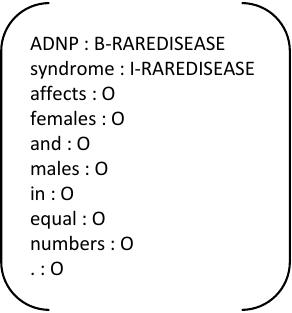}
        \caption{token-level}
        \label{fig:second_subfig}
    \end{subfigure}
    \caption{Different NER outputs for the given text: \textit{ADNP syndrome affects females and males in equal numbers .}}
    \label{fig:ners}
\end{figure}

The difference between document-level and token-level NER is illustrated in Figure~\ref{fig:ners}. 
This gap is significant as token-level NER offers the potential for more detailed and precise entity recognition, which is crucial in clinical applications. For instance, to interpret clinical texts like chemotherapy treatment records, in a scenario where a patient’s record contains multiple chemotherapy sessions over several months, document-level NER may not efficiently distinguish between the different treatment phases and their respective impacts. In contrast, token-level NER can capture specific details. For example, consider a clinical text, ``Patient experienced nausea after chemo session 1 in January and significant fatigue following chemo session 3 in March.'' While document-level NER might identify terms like ``chemo'', ``nausea'', and ``fatigue'', it fails to connect these conditions to specific treatment events. Token-level NER, however, can extract fine-grained details, such as linking ``nausea'' directly to ``chemo session 1 in January'', providing valuable context for understanding and tracking the patient's treatment response over time. This precision is vital for personalized patient care.

This study aims to explore the effectiveness of LLMs in token-level NER for rare diseases within clinical texts. By focusing on this specific application, we seek to understand the limitations and capabilities of LLMs in handling the challenges of token-level entity recognition in a domain where data is scarce and highly specialized. Essentially, we explore the performance of various state-of-the-art local open-source LLMs, including LLaMA-2 \cite{touvron2023llama2}\hspace{-0.1em}, Meditron \cite{chen2023meditron}\hspace{-0.1em}, Llama2-MedTuned \cite{rohanian2023exploring}\hspace{-0.1em}, and UniversalNER \cite{zhou2023universalner}\hspace{-0.1em}, on the task of NER of rare diseases and their phenotypes. Additionally, we assess the performance of prominent models such as ChatGPT-3.5 and ChatGPT-4. Our investigation extends to examining their capabilities in in-context learning, evaluating performance enhancements through few-shot learning, retrieval-augmented generation (RAG), and fine-tuning. The experimental results reveal that without proper fine-tuning, local LLMs generally struggle with token-level NER, despite prior training on clinical texts. We observe that while few-shot learning can effectively improve the performance of most LLMs, the influence of RAG on the same task is relatively minimal. Notably, our findings indicate that a medically-adapted LLaMA-2-7b model, specifically Llama2-MedTuned \cite{rohanian2023exploring}\hspace{-0.1em}, can outperform ChatGPT-4 on this task. This is particularly noteworthy as it achieves these results without having been trained specifically on the rare disease data used in the experiments, and this highlights the potential of fine-tuning local LLMs for specialized applications in clinical NER and other clinical NLP tasks. 


\section*{Related Work}
Recently, researchers have released various LLMs dedicated to the medical field. LLaMA \cite{touvron2023llama}\hspace{-0.1em} and LLaMA-2 \cite{touvron2023llama2}\hspace{-0.1em} are one of the first and most popular open-source LLMs, laying the foundation for extensive research and applications. Meditron \cite{chen2023meditron}\hspace{-0.1em} is a suite of open-source medical LLMs pre-trained on the GAP-Replay corpus, including clinical guidelines, research papers, and general domain data, and surpass the performance of LLaMA-2 \cite{touvron2023llama2}\hspace{-0.1em}, ChatGPT-3.5, and Flan-PaLM \cite{chung2022scaling}\hspace{-0.1em} on multiple medical reasoning benchmarks. 

In parallel, considerable efforts have been devoted to adapting these LLMs to the task of NER through prompting or fine-tuning \cite{touvron2023llama,zhou2023universalner,shyr2024identifying,hu2024improving,li2024ae,li2024improving,lu-etal-2021-parameter-efficient,lu-etal-2022-clinicalt5}\hspace{-0.1em}. For example, Zhou \textit{et al.} instruction-fine-tune the LLaMA model \cite{touvron2023llama}\hspace{-0.1em} using ChatGPT-generated synthetic data for NER from broad-coverage unlabeled web text \cite{zhou2023universalner}\hspace{-0.1em}. Their UniversalNER model shows promising NER performance across multiple domains. However, they only generate document-level outputs, i.e., a list of extracted entities in the given text as shown in Figure~\ref{fig:first_subfig}, without considering their exact span information, which limits their impact and usage in practical scenarios. Hu \textit{et al.} explore the token-level NER capabilities of ChatGPT-3.5 and ChatGPT-4 on two clinical NER tasks by manually crafting specific prompts \cite{hu2024improving}\hspace{-0.1em}. However, they limit their exploration to ChatGPT models and do not investigate local and open-source LLMs. Another study that aligns closely with our research is by Shry \textit{et al.}, who focus on extracting rare diseases and their phenotypes at a document-level by prompting ChatGPT-3.5 \cite{shyr2024identifying}\hspace{-0.1em}.

Unlike these studies, we aim to bridge the gap by exploring the capabilities of both local open-source LLMs and ChatGPT models in the specific context of token-level NER in clinical settings. This approach seeks to understand how these models perform in an area that is not only highly specialized but also characterized by its complexity and data scarcity, particularly in the study of rare diseases.

\begin{table*}[t]
\begin{center}\caption{Statistics of the RareDis dataset.}\label{dataset}
\resizebox{0.5\textwidth}{!}{\begin{tabular}{lllll}
    \toprule
    \bf Data Type & \bf Training & \bf Validation & \bf Test &\bf  Total\\
    \midrule
    Documents   &729    &104    &208    &1041\\
    Sentences   &6281    &894    &1735    &8910\\
    \midrule
    Disease        &1328    &187    &392    &1907\\
    Rare Disease   &3075    &468    &918    &4461\\
    Skin Rare Disease        &442    &42    &143    &627\\
    Sign           &3384    &474    &966    &4824\\
    Symptom        &313    &22    &51    &386\\
    \bottomrule
\end{tabular}}
\end{center}
\end{table*}

\section*{Methods}
\paragraph{Task Overview}
Token-level named entity recognition (NER), also known as span-based NER, involves identifying and classifying named entities within a text into predefined categories such as diseases, symptoms, genes, etc. Formally, in a sentence with $N$ tokens $X=[x_1,x_2,\dots,x_N]$, an entity is defined as a contiguous span of tokens $e=[x_i,\dots,x_j]$, where $0 \leq i \leq j \leq N$, and each is associated with a specific entity type. The core task is treated as a sequence labeling problem, where the goal is to assign a corresponding sequence of labels $Y=[y_1,y_2,\dots,y_N]$ to the sentence $X$. In this study, the BIO (Beginning, Inside, Outside) schema is employed for labeling. According to this scheme, the first token of an entity of a certain \texttt{type} is tagged as \texttt{B-type}, any subsequent tokens within the same entity are tagged as \texttt{I-type}, and tokens not part of an entity are tagged as \texttt{O}. In this regard, the following eleven categories or labels are used in the experiments: \texttt{O}, \texttt{B-Disease}, \texttt{I-Disease}, \texttt{B-RareDisease}, \texttt{I-RareDisease}, \texttt{B-SkinRareDisease}, \texttt{I-SkinRareDisease}, \texttt{B-Symptom}, \texttt{I-Symptom}, \texttt{B-Sign}, and \texttt{I-Sign}.

In this study, we aim to understand the capabilities of various general and medical LLMs in token-level NER, which is underexplored in clinical settings. Similar to Shry \textit{et al.}'s work \cite{shyr2024identifying}\hspace{-0.1em}, we choose rare diseases as our case study. Our focus is particularly on extracting information about rare diseases and their associated phenotypes, driven by two primary reasons: 1) Data related to rare diseases is scarce and highly specialized, presenting an ideal scenario to leverage the strengths of LLMs; and 2) Patients with rare diseases are a relatively understudied group, necessitating increased research and advocacy efforts, as highlighted by Nguyen \textit{et al.} \cite{nguyen2022advocacy}\hspace{-0.1em} To this end, our investigation spans across the overall performance of both local open-source LLMs and ChatGPT models on this task. More specifically, we initially explore the zero-shot performance of these LLMs using manually-designed prompts. Secondly, to enhance the models' performance for this specific task, we employ in-context learning strategies, i.e., few-shot learning and retrieval-augmented generation (RAG). These approaches provide rich information to the prompts to facilitate a deeper understanding of the task by the LLMs. Finally, we instruction-fine-tune Llama2-MedTuned \cite{rohanian2023exploring}\hspace{-0.1em} on the rare disease dataset to assess the performance of LLMs in a fully supervised way. Our intention is to explore how well LLMs, when fine-tuned with direct supervision, can adapt to and accurately identify entities in complex rare disease data.

\paragraph{Dataset}
In our experiments, we use the RareDis-v1 dataset \cite{martinez2022raredis,segura2022exploring}\hspace{-0.1em}, a curated collection of texts from the National Organization for Rare Disorders (NORD) database\footnote{\url{https://rarediseases.org/}}. This dataset specifically comprises selected sections from NORD articles, which have been manually annotated to identify five key entity types: \texttt{Disease}, \texttt{Rare Disease}, \texttt{Skin Rare Disease}, \texttt{Symptom}, and \texttt{Sign}. The statistics of the dataset are detailed in Table~\ref{dataset}, where the first two rows show the number of documents and sentences, respectively. The last four rows are a breakdown of the specific clinical entities present in the dataset. It is important to distinguish between \texttt{Sign}, which are objectively observable indicators or test results suggesting a disease, and \texttt{Symptom}, which are subjective experiences reported by the patient \cite{martinez2022raredis}\hspace{-0.1em}.

\begin{figure}[t]
    \centering
    \begin{subfigure}[b]{0.3\textwidth}
        \includegraphics[width=\textwidth]{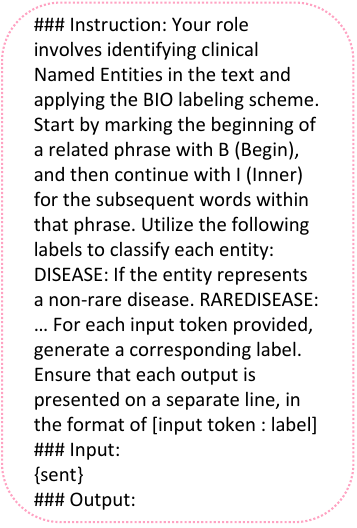}
        \caption{zero-shot}
        \label{fig:first_subfig}
    \end{subfigure}
    \hfill 
    \begin{subfigure}[b]{0.3\textwidth}
        \includegraphics[width=\textwidth]{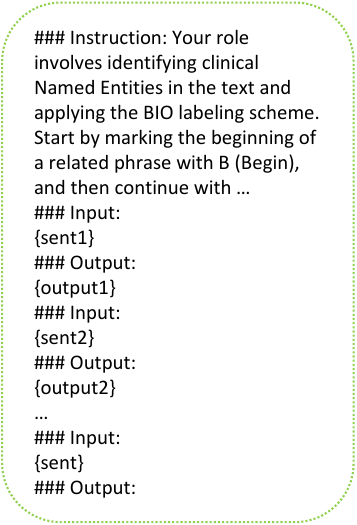}
        \caption{few-shot}
        \label{fig:second_subfig}
    \end{subfigure}
    \hfill 
    \begin{subfigure}[b]{0.3\textwidth}
        \includegraphics[width=\textwidth]{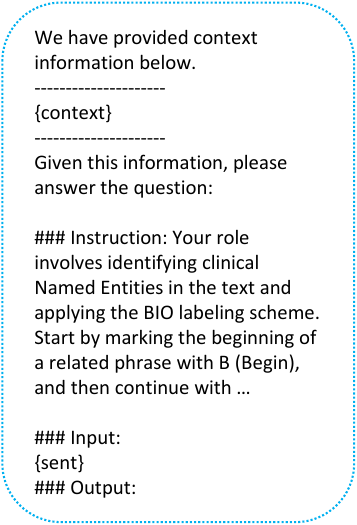}
        \caption{RAG}
        \label{fig:third_subfig}
    \end{subfigure}
    \caption{Prompt templates for zero-shot, few-shot, and RAG-based inference.}
    \label{fig:prompts}
\end{figure}

\paragraph{Models}
We consider the following LLMs for our experiments:
\begin{itemize}
    \item LLaMA-2-7b \cite{touvron2023llama2}\hspace{-0.1em} is one of the first and most popular local open-source pre-trained LLMs released. Though it is not dedicated to the clinical domain, the model is often considered a critical benchmark in recent related studies due to its impact and generalizability. For consistency, we utilize the 7-billion-parameter version of all local LLMs in this study.
    \item Meditron-7b \cite{chen2023meditron}\hspace{-0.1em} is an adapted version of LLaMA-2 to the medical domain through continued pre-training on a comprehensively curated medical corpus, including selected PubMed papers and abstracts, a new dataset of internationally-recognized medical guidelines, and general domain data from RedPajama-v1 \cite{together2023redpajama}\hspace{-0.1em}. It outperforms LLaMA-2-7B on multiple medical reasoning tasks.
    \item UniversalNER-7b \cite{zhou2023universalner}\hspace{-0.1em} is an instruction-fine-tuned version of LLaMA-7b \cite{touvron2023llama}\hspace{-0.1em} which is specified in named entity recognition. Essentially, the authors prompt ChatGPT to generate instruction-tuning data for NER from web text and conduct instruction-tuning on LLaMA. UniversalNER shows state-of-the-art document-level NER performance on multiple benchmarks across diverse domains, including biomedicine. 
    \item Llama2-MedTuned-7b \cite{rohanian2023exploring}\hspace{-0.1em} is a medically-adapted version of the LLaMA-2-7b \cite{touvron2023llama2}\hspace{-0.1em} model. Essentially, the authors specifically create a medical instruction-based dataset consisting of approximately $200,000$ samples covering clinical NLP tasks including token-level named entity recognition, relation extraction, document classification, question answering, natural language inference, and conduct instruction-tuning on LLaMA. Llama2-MedTuned demonstrates on-par performance with domain-specific models like BioClinicalBERT \cite{alsentzer-etal-2019-publicly}\hspace{-0.1em} on several benchmarks.
    \item ChatGPT-3.5 and ChatGPT-4 are arguably by far the most capable LLMs and have demonstrated superior performance and robustness across various domains. However, their proprietary nature limits their use in the clinical domain, where data privacy is of utmost importance. Specifically, we use \texttt{gpt-35-turbo-1106} and \texttt{gpt-4-1106-preview} in the experiments\footnote{We access the OpenAI models through Microsoft Azure's HIPAA-certified platform.}.
\end{itemize}

\paragraph{Learning Strategies}

In this study, we consider both in-context learning and instruction-fine-tuning to further adapt the LLMs to the token-level NER task. For in-context learning, we employ few-shot learning and retrieval-augmented generation (RAG), the prompt templates of which are shown in Figure~\ref{fig:prompts}. Essentially, we randomly select $5$ samples from the training set as the demonstration to the LLMs. To ensure consistency, we fix the $5$ samples for all few-shot experiments in the study. For RAG implementation, we utilize the LlamaIndex framework, incorporating NORD rare disease articles as the knowledge base. Specifically, these articles are segmented into chunks ($chunk\_size=512$), embedded using the \texttt{bge-large-en-v1.5} model, and stored in a vector database. In the retrieval stage, we convert a query using the embedding model and retrieve the top-$k$ ($similarity\_top\_k=2$) most relevant text chunks based on similarity with the query embedding. The retrieved chunks are then integrated into the LLM prompts for enriched context.

For instruction-fine-tuning, we follow previous work \cite{rohanian2023exploring}\hspace{-0.1em} and convert the RareDis dataset to the Stanford Alpaca \cite{alpaca}\hspace{-0.1em} format. This format is structured for instruction-following, comprising three parts: an instruction, an input, and an output, as shown in Figure~\ref{fig:prompts}. We then fine-tune the Llama2-MedTuned-7b model on the transformed RareDis dataset. We use LoRA adapters \cite{hu2022lora}\hspace{-0.1em} for the fine-tuning. It is worth noting that while instruction-fine-tuning (or instruction-tuning) somewhat contradicts our initial intent due to its reliance on sufficient data, it remains crucial for our investigation into LLMs' performance capabilities and limitations in comparison to smaller BERT-like models in data-rich scenarios. Further implementation details are available in our code repository\footnote{\url{https://github.com/qiuhaolu/tner}}.

\begin{table*}[t]
\begin{center}\caption{Zero-shot performance of diverse LLMs on token-level NER.}\label{table2}
\begin{tabular}{lllllllll}
    \toprule
    \multirow{2}{7em}{\bf Model} &\multirow{2}{5em}{\bf Specificity} &\multirow{2}{5em}{\bf Source} & \multicolumn{3}{c}{\bf Disease} & \multicolumn{3}{c}{\bf Rare Disease}\\
    & & & \bf P & \bf R & \bf F1 & \bf P & \bf R & \bf F1\\
    \midrule
    LLaMA-2 & General & Open-Source& 0.0564 & 0.0149 & 0.0219 & 0.3548 & 0.1066 & 0.1494\\
    Meditron & Medical & Open-Source& 0.0586 & 0.0242 & 0.0318 & 0.3409 & 0.1493 & 0.1877\\
    UniversalNER & Task-Specific & Open-Source& 0.1544 & 0.0731 & 0.0936 & 0.5493 & 0.1634 & 0.2267\\
    Llama2-MedTuned & Medical & Open-Source& 0.1391 &\bf 0.7201 & 0.2332 & 0.3988 &\bf 0.8080 & 0.5340\\
    \midrule
    ChatGPT-3.5 & General & Proprietary& 0.1173 & 0.3842 & 0.1798 &\bf 0.5810 & 0.2513 & 0.3509\\
    ChatGPT-4 & General & Proprietary&\bf 0.1993 & 0.4377 &\bf 0.2739 & 0.5767 & 0.5761 &\bf 0.5764\\
    \bottomrule
\end{tabular}
\end{center}
\end{table*}

\paragraph{Evaluation}
We utilize Precision (P), Recall (R), and F1-score (F1) as our evaluation metrics for the token-level NER task. For models designed for document-level NER, such as UniversalNER-7b \cite{zhou2023universalner}\hspace{-0.1em}, we adapt their outputs to token-level by exact string matching. Initially, we assess a range of LLMs to investigate their overall performance. This is followed by a more focused evaluation of those LLMs that demonstrate proficiency in token-level NER. Subsequently, we conduct a detailed analysis of the aforementioned learning strategies, examining their influence on the performance of the LLMs. Additionally, we perform an error analysis to gain insights into the common errors across different models and learning strategies.

\section*{Results}

\paragraph{LLM Initial Evaluation}

In our initial evaluation, we focus on assessing the performance of the aforementioned LLMs in token-level NER using the RareDis-v1 dataset, particularly targeting the two key target entities, i.e., \texttt{Disease} and \texttt{Rare Disease}. The results in Table~\ref{table2} show that, generally, all the models struggle with this task, as evidenced by their modest performance metrics. Notably, Meditron, with its medical training, demonstrates marginally better results than LLaMA-2. However, even UniversalNER, which is optimized for document-level NER, does not show significantly improved performance in this token-level task. In contrast, Llama2-MedTuned-7b demonstrates promising results, closely trailing behind ChatGPT-4, which highlights the potential of local open-source LLMs in clinical NLP applications, specifically in token-level NER.

\begin{table*}[t]
\begin{center}\caption{In-depth performance analysis of ChatGPT and Llama2-MedTuned across different entity types.}\label{table3}
\resizebox{1\textwidth}{!}{\begin{tabular}{lllllllllll}
    \toprule
    \multirow{2}{4em}{\bf Prompting}&\multirow{2}{7em}{\bf Entity Type} & \multicolumn{3}{c}{\bf ChatGPT-3.5}& \multicolumn{3}{c}{\bf ChatGPT-4} & \multicolumn{3}{c}{\bf Llama2-MedTuned}\\
    & & \bf P &\bf  R & \bf  F1 & \bf P & \bf R & \bf  F1 & \bf  P &\bf   R & \bf  F1\\ 
    \midrule
    \multirow{6}{4em}{zero-shot}&Disease & 0.1173 & 0.3842 & 0.1798 & 0.1993 & 0.4377 & 0.2739 & 0.1391 & 0.7201 & 0.2332\\
    &Rare Disease&0.5810&0.2513&0.3509&0.5767&0.5761&0.5764&0.3988&0.8080&0.5340\\
    &Skin Rare Disease&0.1200&0.0833&0.0984&0.4762&0.1389&0.2151&0.0686&0.8542&0.1270\\
    &Sign &0.1314&0.0844&0.1028&0.1802&0.2665&0.2150&0.1712&0.3364&0.2269\\
    &Symptom&0.0305&0.5882&0.0580&0.0574&0.7255&0.1063&0.0123&0.4902&0.0240\\
    \cmidrule{2-11}
    &Average&0.1960&0.2783&0.1580&0.2980&0.4289&0.2773&0.1580&0.6418&0.2290\\
    \midrule
    \multirow{6}{4em}{few-shot}&Disease & 0.1871 & 0.2723 & 0.2218 & 0.2254 & 0.2977 & 0.2566 & 0.1517 & 0.5038 & 0.2332\\
    &Rare Disease&0.5615&0.5512&0.5563&0.6011&0.5901&0.5955&0.5106&0.8069&0.6254\\
    &Skin Rare Disease&0.1231&0.3958&0.1878&0.2953&0.3958&0.3383&0.0823&0.8333&0.1498\\
    &Sign &0.1681&0.0813&0.1096&0.2749&0.2500&0.2619&0.1661&0.3283&0.2206\\
    &Symptom&0.0478&0.6471&0.0889&0.0900&0.5490&0.1547&0.0140&0.5294&0.0273\\
    \cmidrule{2-11}
    &Average&0.2175&0.3895&0.2329&0.2973&0.4165&0.3214&0.1849&0.6003&0.2513\\
    \midrule
    \multirow{6}{4em}{RAG}&Disease & 0.1568 & 0.2952 & 0.2048 & 0.2211 & 0.3995 & 0.2847 & 0.1410 & 0.6412 & 0.2312\\
    &Rare Disease&0.6625&0.4595&0.5427&0.6552&0.5534&0.6000&0.4813&0.7918&0.5987\\
    &Skin Rare Disease&0.1333&0.0833&0.1026&0.7419&0.1597&0.2629&0.0886&0.7986&0.1595\\
    &Sign &0.1586&0.0607&0.0807&0.2263&0.2037&0.2144&0.0955&0.0802&0.0872\\
    &Symptom&0.0420&0.6078&0.0786&0.0482&0.7451&0.0906&0.0173&0.4314&0.0314\\
    \cmidrule{2-11}
    &Average&0.2306&0.3013&0.2033&0.3785&0.4123&0.2905&0.1645&0.5486&0.2216\\
    \bottomrule
\end{tabular}}
\end{center}
\end{table*}

\paragraph{Focused Evaluation of Token-Level NER Capable LLMs}
Table~\ref{table3} provides an in-depth performance analysis of ChatGPT-3.5, ChatGPT-4, and Llama2-MedTuned, selected based on their promising results in our initial evaluation. We encompass a range of experimental settings in this subsection, including zero-shot, few-shot, and retrieval-augmented generation (RAG). The discussion on the fine-tuning approach is reserved for the following subsection, as it requires the use of the entire training dataset. 

A notable observation is the relatively high performance in identifying \texttt{Rare Disease}, likely due to its uniqueness and specificity. However, the models have varying degrees of difficulty in identifying subjective experiences reported by the patient, i.e., \texttt{Symptom}, as reflected by their close-to-zero precision and F1-scores. Moreover, we show that few-shot learning can effectively improve the performance across all models, with an increase ranging from $3\%$ to $8\%$ in average F1 scores. We also observe that ChatGPT-3.5 shows a substantial improvement of $20\%$ in the F1-score for \texttt{Rare Disease}. These observations suggest that few-shot learning serves as an effective method to enhance LLMs' performance in token-level NER.

On the other hand, while RAG shows some promise, its overall impact on enhancing model performance in this task is limited. The data shows that RAG particularly benefits the identification of \texttt{Rare Disease} and \texttt{Skin Rare Disease} across all three models. This improvement could be attributed to RAG's ability to leverage external knowledge bases, i.e., NORD articles about rare diseases, which might be rich in rare disease terminology. However, for more general and common entities like \texttt{Disease}, \texttt{Symptom} and \texttt{Sign}, RAG's advantages are less impressive. This highlights the need for further refinement of RAG methods in clinical NLP applications.

\begin{figure}
    \centering
    \includegraphics[width=1\linewidth]{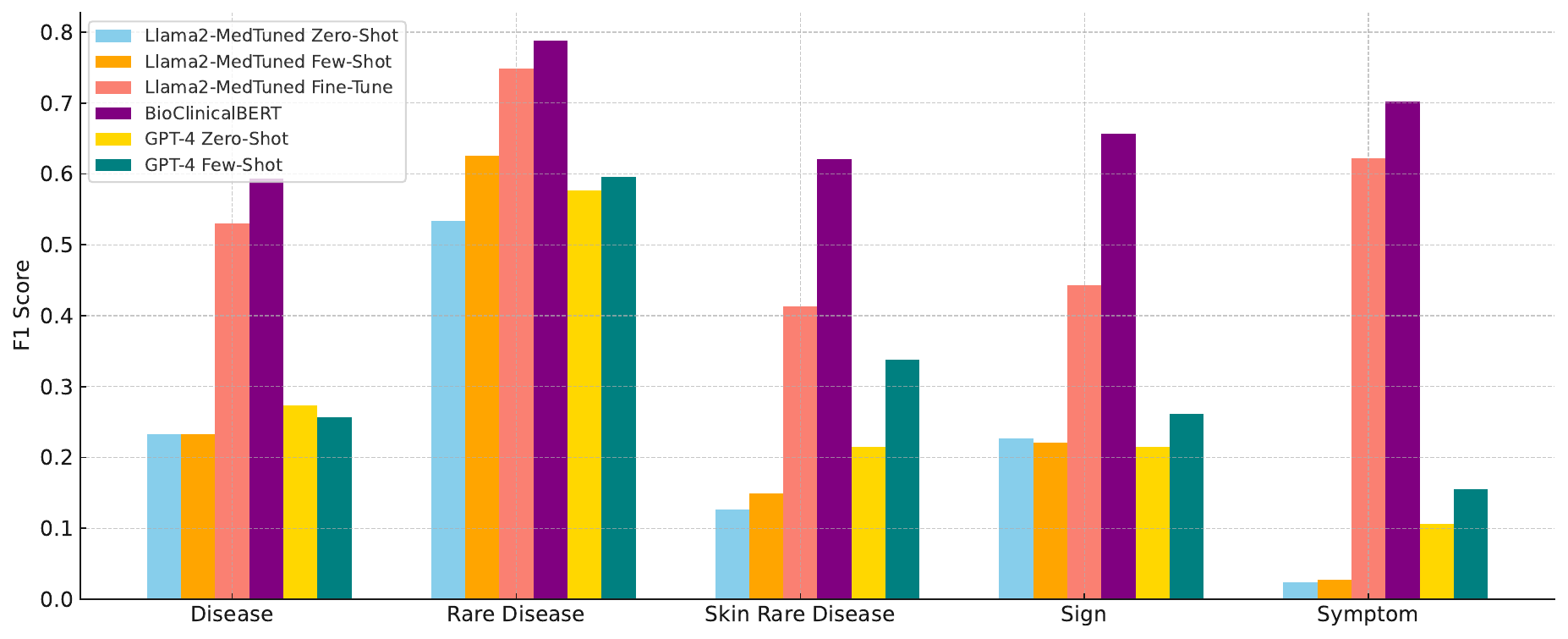}
    \caption{Performance evaluation of Llama2-MedTuned, BioClinicalBERT, and ChatGPT-4 across zero-shot, few-shot, and fine-tuning methods.}
    \label{fig:finetune}
\end{figure}

Besides zero-shot prompting and in-context learning, we also investigate the impact of instruction-fine-tuning on these LLMs in this task. As shown in Table~\ref{table2} and Table~\ref{table3}, despite the performance gain via few-shot learning, the overall performance of prompting LLMs is far from satisfactory. In this experiment, we aim to investigate LLMs' performance capabilities and limitations compared to fine-tuned BERT-like models under data-rich conditions, which is the common solution to NER in current practice. The results are shown in Figure~\ref{fig:finetune}. Essentially, Figure~\ref{fig:finetune} visualizes the F1 scores of various models, including BioClinicalBERT fine-tune, Llama2-MedTuned under different learning strategies (zero-shot, few-shot, fine-tune), and ChatGPT-4 in zero-shot and few-shot scenarios. Specifically, the performance of Llama2-MedTuned models demonstrates significant effectiveness, not only outperforming ChatGPT-4 in most scenarios but also closely rivaling BioClinicalBERT when fine-tuned. This highlights the potential of open-source LLMs in this task.

\paragraph{Error Analysis}
We present an error analysis to further understand the mistakes made by the LLMs in this task. Essentially, we randomly select $50$ sentences from the test dataset and manually categorize the errors for each incorrect prediction. We investigate the two best-performing models, i.e., ChatGPT-4 under few-shot learning and Llama2-MedTuned under fine-tuning. We identify $4$ types of errors in the experiment, i.e., Inaccurate Boundary, where the model incorrectly identifies the start or end of an entity; Wrong Type, where the entity is recognized, but its type is misclassified; False Negative, where the model overlooks an entity present in the text; and False Positive, where the model mistakenly identifies a non-entity as an entity.

\begin{figure}[t]
    \centering
    \begin{subfigure}[b]{0.48\textwidth}
        \includegraphics[width=\textwidth]{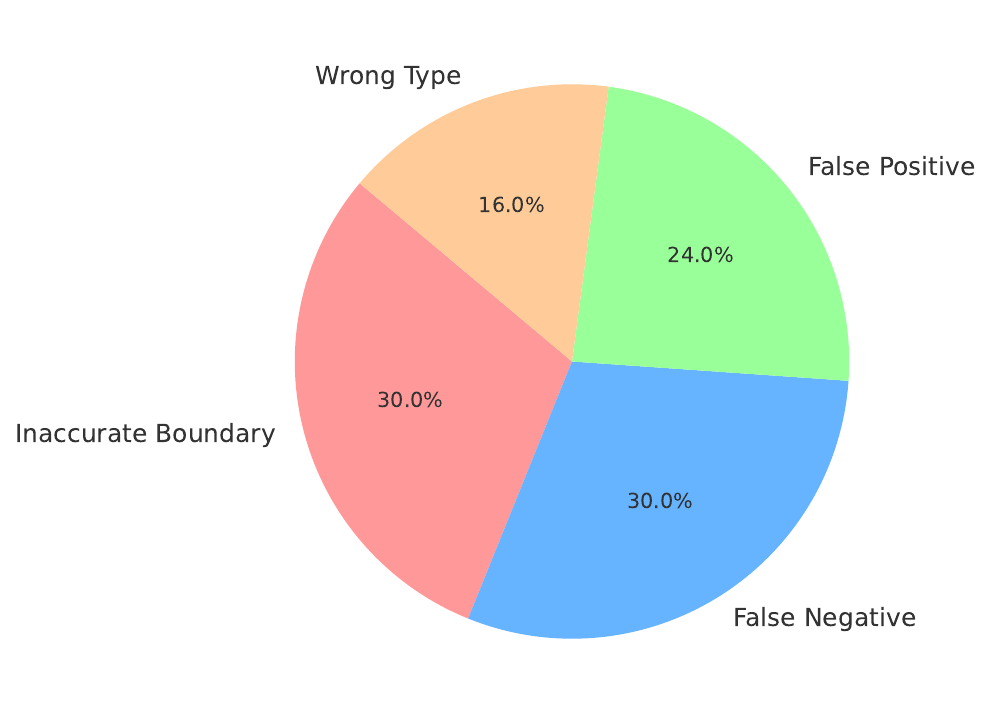}
        \caption{ChatGPT-4}
        \label{figerror:ChatGPT-4}
    \end{subfigure}
    \hfill 
    \begin{subfigure}[b]{0.48\textwidth}
        \includegraphics[width=\textwidth]{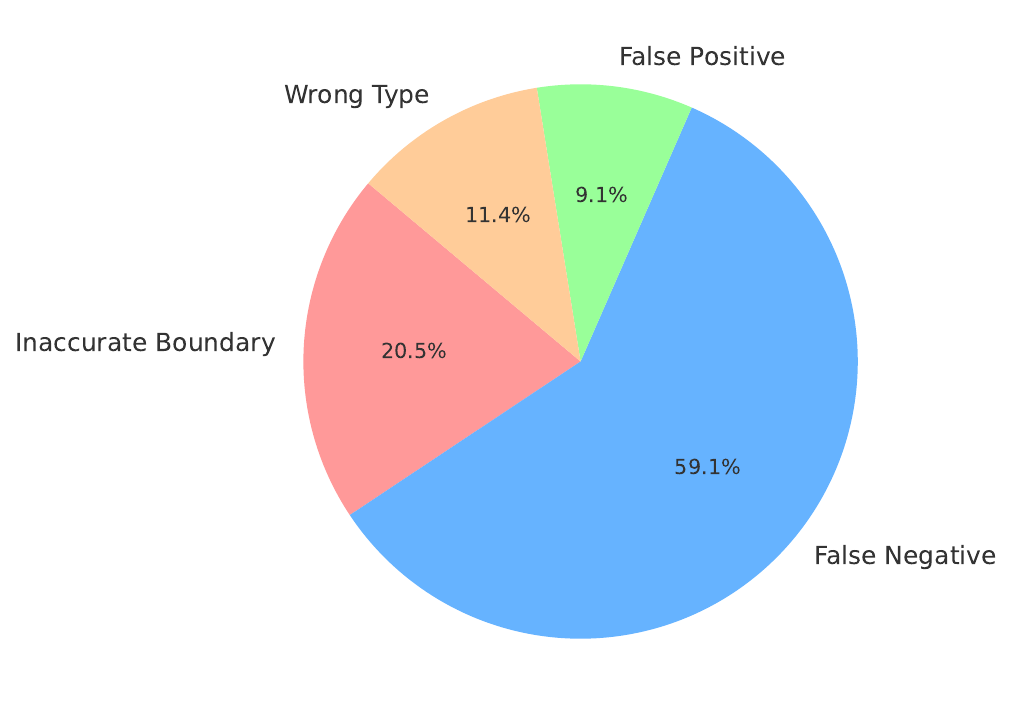}
        \caption{Llama2-MedTuned}
        \label{figerror:Llama2-MedTuned}
    \end{subfigure}
    \caption{Comparative error analysis of ChatGPT-4 and Llama2-MedTuned.}
    \label{fig:error}
\end{figure}

The results in Figure~\ref{fig:error} show that Inaccurate Boundary errors are more prevalent for ChatGPT-4, suggesting challenges in precisely detecting entity boundaries. In contrast, Llama2-MedTuned demonstrates a much higher chance of False Negatives, indicating its inability to identify entities that are present. Both models show fewer Wrong Type and False Positive errors, reflecting a relatively stronger performance in correctly classifying entities and avoiding over-detection. Insights from these error patterns highlight the strengths and weaknesses of each model in token-level NER and indicate directions for future improvements, especially in fine-tuning local open-source LLMs.

\section*{Discussion and Conclusion}


In this study, we investigate the capabilities of LLMs in token-level NER for rare diseases and their associated phenotypes on the RareDis-v1 dataset. Essentially, we find that in general, most LLMs struggle with this challenging task. We also find that a medically-adapted LLaMA-2-7b model, i.e., Llama2-MedTuned, surpasses the performance of ChatGPT-3.5 and matches the performance of ChatGPT-4 on this task, highlighting the potential of local open-source LLMs in clinical NLP applications. Our further analysis suggests that via in-context learning, i.e., few-shot learning and RAG, the performance can be improved while RAG has a relatively limited influence. We also show that after fine-tuning Llama2-MedTuned on this dataset, its performance gets boosted and significantly outperforms that of ChatGPT-4, approaching the levels of BioClinicalBERT. It is worth noting that we use LoRA adapters to perform the fine-tuning, instead of the entire model. Based on these results, we anticipate that full model fine-tuning could potentially enable the Llama2-MedTuned to match BioClinicalBERT's performance.

Large language models (LLMs) have demonstrated superior performance in question-answering-related tasks across almost all domains, including the clinical field. However, their performance in conventional NLP tasks, such as information extraction, has been questioned. Unlike existing studies that either focus on document-level NER or solely rely on the prompt engineering of ChatGPT-series models, our experiment covers a broad scope of LLMs, from proprietary to local open-source and from general-purpose to medically adapted models. We thoroughly evaluate the LLMs on the token-level NER task, showing their struggle and potential directions for refinement. This study could shed light on adapting LLMs to specific NLP applications in clinical settings.

There are several factors that contribute to the struggle of LLMs on this task. Firstly, token-level NER is more challenging than document-level NER. Token-level NER is a token classification task and places significant emphasis on the representation of each individual token. This is in contrast to document-level NER which is a sequence classification task where a pooled representation is often sufficient for predictions, making the role of individual token representation less critical. Secondly, the foundational pre-training objective of most LLMs (decoder-only transformers) is causal language modeling, i.e., next token prediction. Unlike encoder-only transformers like BERT, LLMs don't have an encoder structure that is pre-trained to maximize the model's text representation power, making them less effective in classification tasks. Thirdly, token-level NER often involves a variety of special, previously unseen symbols, such as the diverse BIO tags. These unique elements can present additional complexities for LLMs to understand and operate.

There are also a few options to adapt these LLMs to the token-level NER task. A typical solution is continuous pre-training, or instruction-fine-tuning, just as Llama2-MedTuned and what we do using the rare disease data. Basically, the idea is to cast the NER problem as text generation, so the LLM can have a better understanding of the task and thus process it correctly in the same manner as it was pre-trained. For instance, Yang \textit{et al.} integrate Human Phenotype Ontology (HPO) labels into clinical notes and fine-tune GPT-based models (e.g., GPT-J) for phenotype recognition \cite{yang2024enhancing}\hspace{-0.1em}. They also conduct a comprehensive comparative analysis of encoder-based and decoder-based models for this task. Another solution is to leverage LLMs to generate synthetic data and use the data to train smaller specified models. In addition to the data, one can also alter the model architecture of LLMs. For example, Li \textit{et al.} propose to fine-tune LLaMA-2 models in a label-supervised way, by removing the causal mask from the decoders to enable bidirectional self-attention which is essential for token classification tasks like NER \cite{li2023label}\hspace{-0.1em}. However, in our preliminary experiment, the method does not work as expected and the performance is worse than instruction-fine-tuning on the rare disease dataset. We leave it for future study.

This study has certain limitations. Firstly, it does not explore the potential benefits of specific prompt engineering techniques, such as utilizing feedback from error analysis to refine prompts and thereby enhance model performance. Additionally, our approach mainly focuses on prompting LLMs to generate BIO tags. However, there are alternative approaches that are unexplored, such as the generation of text marked by special symbols to represent span information, which could offer a different perspective on model efficacy. Furthermore, in terms of instruction-fine-tuning of LLMs, our methodology is limited to employing LoRA adapters. A more extensive approach such as full model fine-tuning remains untested and could be valuable for future research.

To conclude, in this study, we explore the capabilities and limitations of existing LLMs, especially local open-source LLMs, in the task of token-level NER of rare diseases and their associated phenotypes. Through this exploration, we aim to contribute to the advancement of LLM-based token-level NER methodologies in the clinical domain, ultimately aiding in the improvement of patient care by enhancing the processing and understanding of clinical texts.

\subparagraph{Acknowledgments}
This work was conducted under support from the National Human Genome Research Institute (R01HG12748).

\makeatletter
\renewcommand{\@biblabel}[1]{\hfill #1.}
\makeatother

\bibliographystyle{vancouver}
\bibliography{amia}  

\end{document}